\documentclass[runningheads]{llncs}
\usepackage{graphicx}
\usepackage{times}
\usepackage{amsmath,amssymb} 
\usepackage{color}
\usepackage{wrapfig}
\usepackage{booktabs}
\usepackage{caption}

\usepackage[width=122mm,left=12mm,paperwidth=146mm,height=193mm,top=12mm,paperheight=217mm]{geometry}
\usepackage{lib}

\usepackage[colorlinks]{hyperref}

\usepackage{floatrow}
\newfloatcommand{capbtabbox}{table}[][\FBwidth]

\begin{document}
\pagestyle{headings}
\newcommand\blfootnote[1]{%
  \begingroup
  \renewcommand\thefootnote{}\footnote{#1}%
  \addtocounter{footnote}{-1}%
  \endgroup
}
\mainmatter
\title{Learning Category-Specific Mesh Reconstruction \\ from Image Collections}
\titlerunning{Category-Specific Mesh Reconstruction}
\author{Angjoo Kanazawa$^*$, Shubham Tulsiani$^*$, Alexei A. Efros, Jitendra Malik}
\institute{University of California, Berkeley\\
{\tt\small\{kanazawa,shubhtuls,efros,malik\}@eecs.berkeley.edu}}
\authorrunning{A. Kanazawa$^*$, S. Tulsiani$^*$, A. A. Efros, J. Malik }

\maketitle
\begin{abstract}
We present a learning framework for recovering the 3D shape, camera, and texture
of an object from a single image. The shape is represented as a deformable 3D
mesh model of an object category where a shape is parameterized by a learned
mean shape and per-instance predicted deformation. Our approach allows
leveraging an annotated image collection for training, where the deformable
model and the 3D prediction mechanism are learned without relying on
ground-truth 3D or multi-view supervision. Our representation enables us to go
beyond existing 3D prediction approaches by  incorporating texture inference as
prediction of an image in a canonical appearance space. Additionally, we show
that semantic keypoints can be easily associated with the predicted shapes. We
present qualitative and quantitative results of our approach on CUB and PASCAL3D
datasets and show that we can learn to predict diverse shapes and textures across
objects using only annotated image collections. The project website can be found at {\footnotesize \url{https://akanazawa.github.io/cmr/}}.
\blfootnote{$^*$ The first two authors procrastinated equally on this work.}
\end{abstract}
\begin{figure*}[h]
  \centering
  \includegraphics[width=\textwidth]{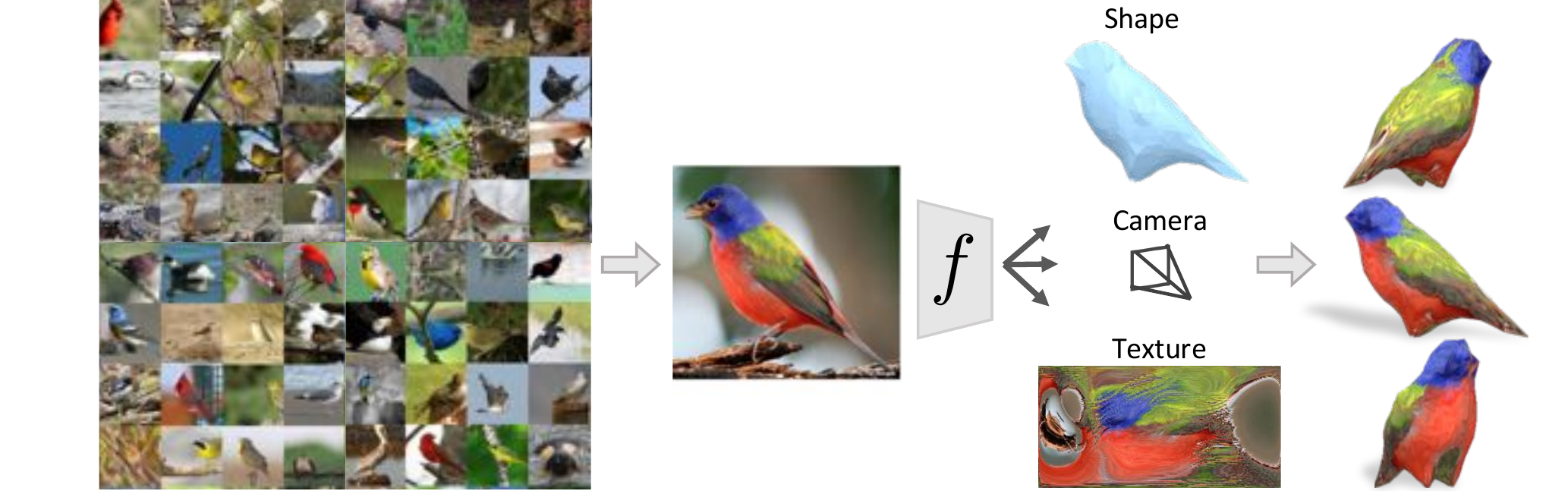}
  \caption{\small Given an annotated image collection of an object category, we
    learn a predictor $f$ that can map a novel image $I$ to its 3D shape, camera
    pose, and texture.}
  \figlabel{teaser}
\end{figure*}

\section{Introduction}
Consider the image of the bird in \figref{teaser}. 
Even though this flat two-dimensional picture printed on a page may be the first time we are seeing this particular bird, we can infer its rough 3D shape, understand the camera pose, and even guess what it would look like from another view. We can do this because all the previously seen birds have enabled us to develop a mental model of what birds are like, and this knowledge helps us to recover the 3D structure of this novel instance. 

In this work, we present a computational model that can similarly learn to infer a 3D representation given just a single image.  As illustrated in \figref{teaser}, the learning only relies on an annotated 2D image collection of a given object category, comprising of foreground masks and semantic keypoint labels. Our training procedure, depicted in \figref{overview}, forces a common prediction  model to explain all the image evidences across many examples of an object category. This allows us to learn a meaningful 3D structure despite only using a single-view per training instance, without relying on any ground-truth 3D data for learning. 

At inference, given a single unannotated image of a novel instance, our learned model allows us to infer the shape, camera pose, and texture of the underlying object. We represent the shape as a 3D mesh in a canonical frame, where the predicted camera transforms the mesh from this canonical space to the image coordinates. The particular shape of each instance is instantiated by deforming a learned category-specific mean shape with instance-specific predicted deformations. The use of this shared 3D space affords numerous advantages as it implicitly enforces correspondences across 3D representations of different instances. As we detail in \secref{approach}, this allows us to formulate the task of inferring mesh texture of different objects as that of predicting pixel values in a common texture representation. Furthermore, we can also easily associate semantic keypoints with the predicted 3D shapes.

Our shape representation is an instantiation of deformable models, the history of which can be traced back to D'Arcy Thompson~\cite{Thompson}, who in turn was inspired by the work of D\"urer~\cite{Durer}. Thompson observed that shapes of objects of the same category may be aligned through geometrical transformations. Cootes and Taylor \cite{cootes1992active} operationalized this idea to learn a class-specific model of deformation for 2D images. Pioneering work of Blanz and Vetter \cite{BlanzVetter} extended these ideas to 3D shapes to model the space of faces. These techniques have since been applied to model human bodies \cite{Anguelov:2005,SMPL}, hands \cite{taylor2014user,khamis2015learning}, and more recently on quadruped animals \cite{zuffi2017}. Unfortunately, all of these approaches require a large collection of 3D data to learn the model, preventing their application to categories where such data collection is impractical. In contrast, our approach is able to learn using only an annotated image collection.

Sharing our motivation for relaxing the requirement of 3D data to learn morphable models, some related approaches have examined the use of similarly annotated image collections. Cashman and Fitzgibbon \cite{Cashman} use keypoint correspondences and segmentation masks to learn a morphable model of dolphins from images. Kar \etal~\cite{CSDM} extend this approach to general rigid object categories. Both approaches follow a \emph{fitting-based} inference procedure, which relies on mask (and optionally keypoint) annotations at test-time and is computationally inefficient. We instead follow a \emph{prediction-based} inference approach, and learn a parametrized predictor which can directly infer the 3D structure from an unannotated image.
Moreover, unlike these approaches, we also address the task of texture prediction which cannot be easily incorporated with these methods. 

While deformable models have been a common representation for 3D inference, the recent advent of deep learning based prediction approaches has resulted in a plethora of alternate representations being explored using varying forms of supervision. Relying on ground-truth 3D supervision (using synthetic data), some approaches have examined learning voxel~\cite{choy20163d,girdhar16b,zhu17reproj,marrnet}, point cloud~\cite{psgn} or octree~\cite{hsp,octreepred} prediction. While some learning based methods do pursue mesh prediction~\cite{hmr,deformnet,laine2017production,sinha2017surfnet}, they also rely on 3D supervision which is only available for restricted classes or in a synthetic setting. 
Reducing the supervision to multi-view masks~\cite{yan2016perspective,rezende2016unsupervised,drcTulsiani17,gwak2017weakly} or depth images~\cite{drcTulsiani17} has been explored for voxel prediction, but the requirement of multiple views per instance is still restrictive. While these approaches show promising results, they rely on stronger supervision (ground-truth 3D or multi-view) compared to our approach.

\vspace{1mm}
In the context of these previous approaches, the proposed approach differs primarily in three aspects:
\begin{itemize}
     \item \emph{Shape representation and inference method.} We combine the benefits of the classically used deformable mesh representations with those of a learning based prediction mechanism.
     The use of a deformable mesh based representation affords several advantages such as memory efficiency, surface-level reasoning and correspondence association. Using a learned prediction model allows efficient inference from a single unannotated image
     \vspace{1mm} \item \emph{Learning from an image collection.} Unlike recent CNN based 3D prediction methods which require either ground-truth 3D or multi-view supervision, we only rely on an annotated image collection, with only one available view per training instance, to learn our prediction model. 
     \vspace{1mm} \item \emph{Ability to infer texture.}
     There is little past work on predicting the 3D shape and the texture of objects from a single image. Recent \emph{prediction-based} learning methods use representations that are not amenable to textures (\eg voxels). The classical deformable model \emph{fitting-based} approaches cannot easily incorporate texture for generic objects. An exception is texture inference on human faces \cite{BlanzVetter,saito2017photorealistic,sela2017unrestricted,tewari2017mofa}, but these approaches require a large-set of 3D ground truth data with high quality texture maps.  Our approach enables us to pursue the task of texture inference from image collections alone, and we address the related technical challenges regarding its incorporation in a learning framework.
   \end{itemize}

\section{Approach}
\seclabel{approach}
We aim to learn a predictor $f_\theta$ (parameterized as a CNN) that can infer
the 3D structure of the underlying object instance from a single image $I$. The prediction $f_\theta(I)$
is comprised of the 3D shape of the object in a canonical frame, the associated
texture, as well as the camera pose.
The shape representation we pursue in this work is of the form of a 3D
mesh. This representation affords several advantages over alternates like
probabilistic volumetric grids \eg amenability to texturing, correspondence
inference, surface level reasoning and interpretability.

The overview of the proposed framework is illustrated in \figref{overview}. The input image is passed through an encoder to a latent representation that is shared by three modules that estimate the camera pose, shape deformation, and texture parameters. The deformation is added to the learned category-level mean shape to obtain the final predicted shape. The objective of the network is to minimize the corresponding losses when the shape is rendered onto the image. We train a separate model for each object category.

\begin{figure}[t]
  \centering
  \includegraphics[width=\textwidth]{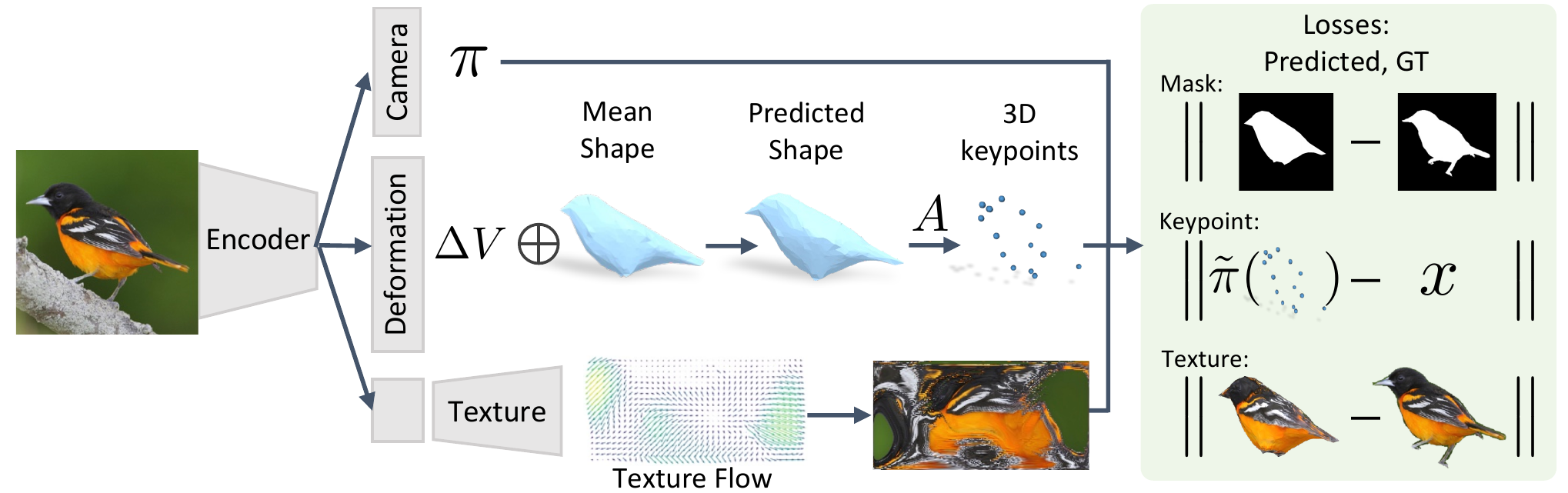}
  \caption{\small {\bf Overview of the proposed framework.} An image $I$ is passed
    through a convolutional encoder to a latent representation that is shared by modules that estimate the camera pose, deformation and texture parameters. Deformation is an offset to the learned mean shape, which when added yield instance specific shapes in a canonical coordinate frame. We also learn correspondences between the mesh vertices and the semantic keypoints. Texture is parameterized as an UV image, which we predict through texture flow (see \secref{texturepred}). The objective is to minimize the distance between the rendered mask, keypoints and textured rendering with the corresponding ground truth annotations. We do not require ground truth 3D shapes or multi-view cues for training.}
  \figlabel{overview}
\end{figure}

We first present the representations predicted by our model in \secref{shaperep}, and then describe the learning procedure in \secref{learning}. We initially present
our framework for predicting shape and camera pose, and then describe how the
model is extended to predict the associated texture in \secref{texturepred}.

\subsection{Inferred 3D Representation}
\seclabel{shaperep}
Given an image $I$ of an instance, we predict $f_{\theta}(I) \equiv (M, \pi)$, a mesh $M$ and camera pose $\pi$ to capture the 3D structure of the underlying object. In addition to these directly predicted aspects, we also learn the association between the mesh vertices and the category-level semantic keypoints. We describe the details of the inferred representations below.

\noindent \paragraph{\emph{\textbf{Shape Parametrization.}}}
We represent the shape as a 3D mesh $M \equiv (V, F)$, defined by vertices $V \in \mathbb{R}^{|V|\times 3}$ and faces $F$. We assume a fixed and pre-determined mesh connectivity, and use the faces $F$ corresponding to a spherical mesh. The vertex positions $V$ are instantiated using
(learned) instance-independent mean vertex locations $\bar{V}$ and
instance-dependent predicted deformations $\Delta_{V}$, which when added, yield
instance vertex locations $V = \bar{V} + \Delta_{V}$. Intuitively, the mean shape $\bar{V}$ can be considered as a learnt bias term for the predicted shape $V$.

\noindent \paragraph{\emph{\textbf{Camera Projection.}}}
We model the camera with weak-perspective projection
and predict, from the input image $I$, the scale $s \in \mathbb{R}$,
translation $\textbf{t} \in \mathbb{R}^2$, and rotation (captured by quaternion $\textbf{q} \in \mathbb{R}^4$). We use $\pi(P)$ to denote the projection of a set of 3D points $P$ onto the image coordinates via the weak-perspective projection defined by $\pi \equiv (s, \textbf{t}, \textbf{q})$.

\noindent \paragraph{\emph{\textbf{Associating Semantic Correspondences.}}}
As we represent the shape using a category-specific mesh in the canonical frame, the regularities across instances encourage semantically consistent vertex positions across instances, thereby implicitly endowing semantics to these vertices. We can use this insight and learn to explicitly associate semantic keypoints \eg, beak, legs \etc with the mesh via a keypoint assignment matrix $A \in \mathcal{R_+}^{|K| \times |V|}$ s.t. $\sum_{v}~A_{k,v}  = 1$. Here, each row $A_k$ represents a probability distribution over the mesh vertices of corresponding to keypoint $k$, and can be understood as approximating a one-hot vector of vertex selection for each keypoint. As we describe later in our learning formulation, we encourage each $A_k$ to be a peaked distribution. Given the vertex positions $V$, we can infer the location $v_k$ for the  $k^{th}$ keypoint as $v_k = \sum_{v} A_{k,v} v$. More concisely, the keypoint locations induced by vertices $V$ can be obtained as $A \cdot V$. We initialize the keypoint assignment matrix $A$ uniformly, but over the course of training it learns to better associate semantic keypoints with appropriate mesh vertices.

In summary, given an image $I$ of an instance, we predict the corresponding camera $\pi$ and the shape deformation $\Delta_{V}$ as $(\pi, \Delta_{V}) = f(I)$. In addition, \emph{we also learn} (across the dataset), instance-independent parameters $\{\bar{V}, A\}$. As described above, these category-level (learned) parameters, in conjunction with the instances-specific predictions, allow us to recover the mesh vertex locations $V$ and coordinates of semantic keypoints  $A \cdot V$.

\subsection{Learning from an Image Collection}
\seclabel{learning}
We present an approach to train $f_\theta$ without relying on strong supervision in the form of ground truth 3D
shapes or multi-view images of an object instance. 
Instead, we guide the
learning from an image collection annotated with sparse keypoints and
segmentation masks. Such a setting is more natural and easily obtained, particularly for animate and deformable objects such as birds or animals. It is extremely difficult to obtain scans, or even multiple views of the same instance for these classes, but relatively easier to acquire a single image for numerous instances.

Given the annotated image collection, we train $f_\theta$ by formulating an objective
function that consists of instance specific losses and priors. The instance-specific energy terms ensure that the predicted 3D structure is consistent with the available evidence (masks and keypoints) and the priors encourage generic desired properties \eg smoothness. As we learn a common prediction model $f_{\theta}$ across many instances, the common structure across the category allows us to learn meaningful 3D prediction despite only having a single-view per instance.

\noindent \paragraph{\emph{\textbf{Training Data.}}}
We assume an annotated training set $\{(I_i, S_i, x_i)\}_{i=1}^N$ for each object category, where $I_i$ is the image, $S_i$ is the instance
segmentation, and $x_i\in \mathbb{R}^{2\times K}$ is the set of  $K$ keypoint
locations. As previously leveraged by ~\cite{Vincente,CSDM}, applying structure-from-motion to the annotated keypoint locations additionally allows us to obtain a rough estimate of the weak-perspective camera $\tilde{\pi}_i$ for each training instance. This results in an augmented training set $\{(I_i, S_i, x_i, \tilde{\pi}_i)\}_{i=1}^N$, which we use for training our predictor $f_{\theta}$.

\noindent \paragraph{\emph{\textbf{Instance Specific Losses.}}}
We ensure that the predicted 3D structure matches the available annotations. Using the semantic correspondences associated to the mesh via the keypoint assignment matrix $A$, we formulate a keypoint reprojection loss. This term encourages the predicted 3D keypoints to match the annotated 2D keypoints when projected onto the image:
\begin{equation}
  \label{eq:reproj}
  L_{\texttt{reproj}} = \sum_i|| x_i - \tilde{\pi}_i(AV_i) ||_2.
\end{equation}
Similarly, we enforce that the predicted 3D mesh, when rendered in the image coordinates, is consistent with the annotated foreground mask: $L_{\texttt{mask}} = \sum_i|| S_i - \mathcal{R}(V_i, F, \tilde{\pi}_i) ||_2$.
Here, $\mathcal{R}(V, F, \pi)$ denotes a rendering of the segmentation mask image corresponding to the 3D mesh $M=(V, F)$ when rendered through camera $\pi$. In all of our experiments, we use Neural Mesh Renderer \cite{NMR} to provide a differentiable implementation of $\mathcal{R}(\cdot)$.

We also train the predicted camera pose to match the corresponding estimate obtained via structure-from-motion using a regression loss $L_{\texttt{cam}} = \sum_i||\tilde{\pi}_i - \pi_i||_2 $. We found it advantageous to use the structure-from-motion camera $\tilde{\pi}_i$, and not the predicted camera $\pi_i$, to define $L_{\texttt{mask}}$ and $L_{\texttt{reproj}}$ losses. This is because during training, in particular the initial stages when the predictions are often incorrect, an error in the predicted camera can lead to high errors despite accurate shape, and possibly adversely affect learning.

\noindent \paragraph{\emph{\textbf{Priors.}}} In addition to the data-dependent losses which ensure that the predictions match the evidence, we leverage generic priors to encourage additional properties. The prior terms that we use are:

\noindent \emph{Smoothness.}
In the natural world, shapes tend to have a smooth surface and we would like our
recovered 3D shapes to behave similarly. An advantage of using a mesh
representation is that it naturally affords reasoning at the surface level. In particular, enforcing smooth surface has been extensively studied
by the Computer Graphics community \cite{pinkall1993computing,sorkine2004laplacian}. Following
the literature, we formulate surface smoothness as minimization of the mean
curvature. On meshes, this is captured by the norm of the graph
Laplacian, and can be concisely written as $L_{\texttt{smooth}} = ||LV||_2$,
where $L$ is the discrete Laplace-Beltrami operator. We construct $L$ once using the
connectivity of the mesh and this can be expressed as a simple linear operator on
vertex locations. See appendix for details.

\noindent \emph{Deformation Regularization.} In keeping with a common practice across deformable model approaches~\cite{BlanzVetter,Cashman,CSDM}, we find it beneficial to regularize the deformations as it discourages arbitrarily large deformations and helps learn a meaningful mean shape. The corresponding energy term is expressed as $L_{\texttt{def}} = ||\Delta_V||_2$.

\noindent \emph{Keypoint association.} As discussed in \secref{shaperep}, we encourage the keypoint assignment matrix $A$ to be a peaked distribution as it should intuitively correspond to a one-hot vector. We therefore minimize the average entropy over all
keypoints: $L_{\texttt{vert2kp}} = \frac{1}{|K|}\sum_{k}\sum_{v}-A_{k,v}\log A_{k,v}$.

In summary, the overall objective for shape and camera is
\begin{equation}
  \eqlabel{loss_overview}
  L =  L_{\texttt{reproj}} + L_{\texttt{mask}} +
  L_{\texttt{cam}} + L_{\texttt{smooth}} + L_{\texttt{def}} + L_{\texttt{vert2kp}}.
\end{equation}

\noindent \paragraph{\emph{\textbf{Symmetry Constraints.}}}
Almost all common object categories, including the ones we consider, exhibit reflectional symmetry. To exploit this structure, we constrain the predicted shape and deformations to be mirror-symmetric. As our mesh topology corresponds to that of a sphere, we identify symmetric vertex pairs in the initial topology. Given these pairs, we only learn/predict parameters for one vertex in each pair for the mean shape $\bar{V}$ and deformations $\Delta_V$. See appendix for details. 

\noindent \paragraph{\emph{\textbf{Initialization and Implementation Details.}}} While our mesh topology corresponds to a sphere, following previous fitting based deformable model approaches~\cite{CSDM}, we observe that a better initialization of the mean vertex positions $\bar{V}$ speeds up learning. We compute the convex hull of the mean keypoint locations obtained during structure-from-motion and initialize the mean vertex locations to lie on this convex hull -- the procedure is described in more detail in the appendix. As the different energy terms in \eqref{loss_overview} have naturally different magnitudes, we weight them accordingly to normalize their contribution.

\subsection{Incorporating Texture Prediction}
\seclabel{texturepred}
\setlength{\columnsep}{10pt}
\setlength{\intextsep}{0pt}
\begin{wrapfigure}{r}{0.5\textwidth}
  \begin{center}
  \includegraphics[width=.9\textwidth]{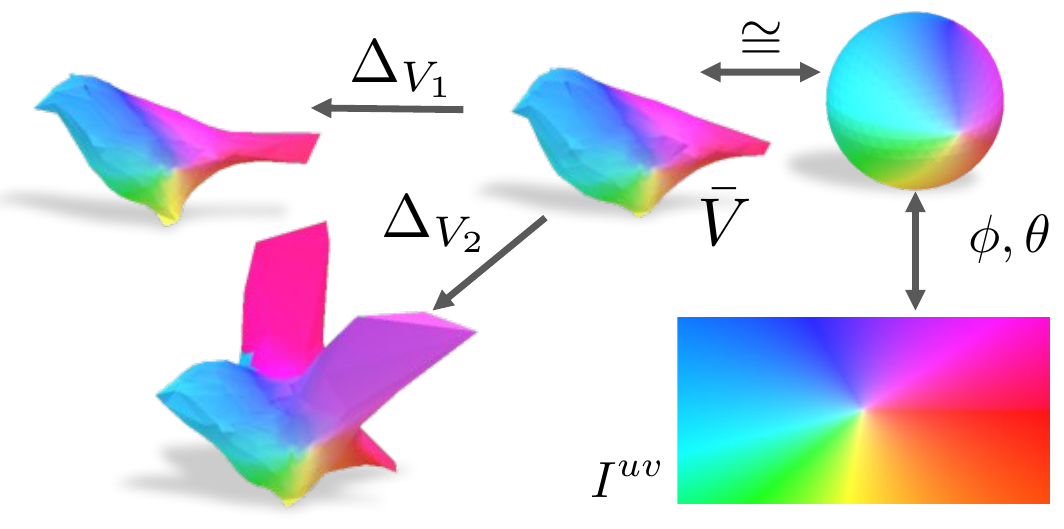}
\end{center}
\setlength{\intextsep}{0pt}
\caption{\small {\bf Illustration of the UV mapping.} We illustrate how a
  texture image $I^{uv}$ can induce a corresponding texture on the predicted
  meshes. A point on a sphere can be mapped onto the image $I^{uv}$ via using
  spherical coordinates. As our mean shape has the same mesh geometry (vertex
  connectivity) as a sphere we can transfer this mapping onto the mean shape. The different predicted shapes, in turn, are simply deformations of the mean shape and can use the same mapping.}
\figlabel{uv_map}
\end{wrapfigure}
In our formulation, all recovered shapes share a common underlying 3D mesh
structure -- each shape is a deformation of the mean shape. We can leverage this
property to reduce texturing of a particular instance to predicting the texture
of the mean shape. Our mean shape is isomorphic to a sphere, whose texture can
be represented as an image $I^{uv}$, the values of which get mapped onto the
surface via a fixed UV mapping (akin to unrolling a globe into a flat map)
\cite{hughes2014computer}.  Therefore, we formulate the task of texture
prediction as that of inferring the pixel values of $I^{uv}$. This image can be
thought of as a canonical appearance space of the object category. For example,
a particular triangle on the predicted shape always maps to a particular region
in $I^{uv}$, irrespective of how it was deformed. This is illustrated in
\figref{uv_map}. In this texture parameterization, each pixel in the UV image
has a consistent semantic meaning, thereby making it easier for the prediction
model to leverage common patterns such as correlation between the bird back and
the body color.

\begin{figure*}[t]
  \centering
\includegraphics[width=\textwidth]{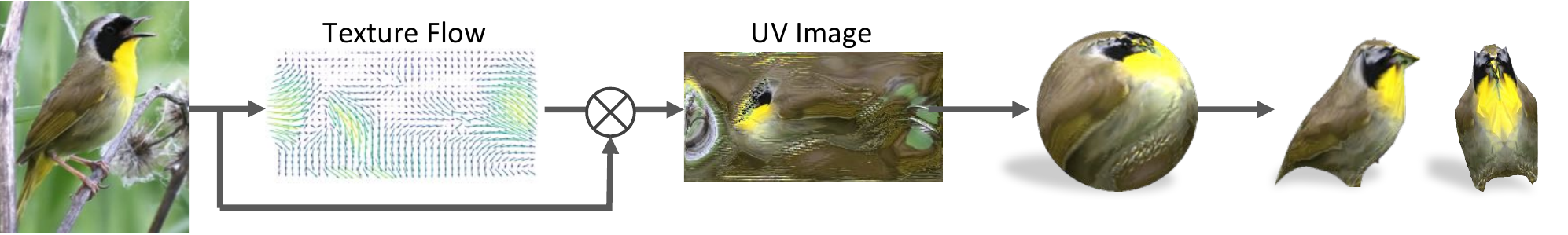}
  \caption{\small {\bf Illustration of texture flow.} We predict a texture flow $\mathcal{F}$ that is used to bilinearly sample the input image $I$ to generate the texture image $I^{uv}$. We can use this predicted UV image $I^{uv}$ to then texture the instance mesh via the UV mapping procedure illustrated in \figref{uv_map}.}
  \figlabel{texture_flow}
\end{figure*}

We incorporate texture prediction module into our framework by setting up a decoder that upconvolves the latent representation to the spatial dimension of $I^{uv}$. 
While directly regressing the pixel values of $I^{uv}$ is a feasible approach, this often results in blurry images. Instead, we take inspiration from \cite{appFlowZhou16} and formulate this task as that of predicting the appearance flow. Instead of regressing the pixel values of $I^{uv}$, the texture module outputs where to copy the color of the pixel from the original input image. This prediction mechanism, depicted in \figref{texture_flow}, easily allows our predicted texture to retain the details present in the input image. We refer to this output as `texture flow' $\mathcal{F} \in \mathbb{R}^{H_{uv} \times W_{uv} \times 2}$, where $H_{uv}, W_{uv}$ are the height and width of $I^{uv}$, and $\mathcal{F}(u,v)$ indicates the $(x,y)$ coordinates of the input image to sample the pixel value from. This allows us to generate the UV image $I^{uv}=G(I; \mathcal{F})$ by bilinear sampling $G$ of the original input image $I$ according to the predicted flow $\mathcal{F}$. This is illustrated in \figref{texture_flow}.

Now we formulate our texture loss, which encourages the rendered texture image to match the foreground image:
\begin{equation}
    \label{eq:texture}
  L_{\texttt{texture}} = \sum_i \textrm{dist}(S_i \odot I_i, S_i \odot \mathcal{R}(V_i, F, \tilde{\pi}_i, I^{uv})).
\end{equation}
$\mathcal{R}(V_i, F, \tilde{\pi}_i, I^{uv}_i)$ is the rendering of the 3D mesh with texture defined by $I^{uv}$. We use the perceptual metric of Zhang \etal \cite{zhang2018perceptual} as the distance metric.

The loss function above provides supervisory signals to regions of $I^{uv}$ corresponding to the foreground portion of the image, but not to other regions of $I^{uv}$ corresponding to parts that are not directly visible in the image. While the common patterns across the dataset \eg similar colors for bird body and back can still allow meaningful prediction, we find it helpful to add a further loss that encourages the texture flow to select pixels only from the foreground region in the image. This can be simply expressed by sampling the distance transform field of the foreground mask
$\mathcal{D}_S$ (where for all points $x$ in the foreground, $\mathcal{D}_S(x) = 0$) according to $\mathcal{F}$ and summing the resulting image:
\begin{equation}
  \label{eq:tex_dt}
  L_{\texttt{dt}} = \sum_{i} \sum_{u,v}G(\mathcal{D}_{S_i}; \mathcal{F}_i)(u,v).
\end{equation}
In contrast to inferring the full texture map, directly sampling the actual pixel values that the predicted mesh projects onto creates holes and leaking of the background texture at the boundaries. Similarly to the shape parametrization, we also explicitly encode symmetry in our $I^{uv}$ prediction, where symmetric faces gets mapped on to the same UV coordinate in $I^{uv}$. Additionally, we only back-propagate gradients from $L_{\texttt{texture}}$ to the predicted texture (and not the predicted shape) since bilinear sampling often results in high-frequency gradients that destabilize shape learning. Our shape prediction is therefore learned only using the objective in \eqref{loss_overview}, and the losses $L_{\texttt{texture}}$ and $L_{\texttt{dt}}$ can be viewed as encouraging prediction of correct texture `on top' of the learned shape.

\section{Experiments}
We demonstrate the ability of our presented approach to learn single-view inference of shape, texture and camera pose using only a category-level annotated image collection. As a running example, we consider the `bird' object category as it represents a challenging scenario that has not been addressed via previous approaches. We first present, in \secref{expsetup}, our experimental setup, describing the annotated image collection and CNN architecture used.

As ground-truth 3D is not available for benchmarking, we present extensive qualitative results in \secref{qualitative}, demonstrating that we learn to predict meaningful shapes and textures across birds. We also show we capture the shape deformation space of the category and that the implicit correspondences in the deformable model allow us to have applications like texture transfer across instances.

We also present some quantitative results to provide evidence for the accuracy of our shape and camera estimates in \secref{quantitative}. While there has been little work for reconstructing categories like birds, some approaches have examined the task of learning shape prediction using an annotated image collection for some rigid classes. In \secref{pascal3d} we present our method's results on some additional representative categories, and show that our method performs comparably, if not better than the previously proposed alternates while having several additional advantages \eg learning semantic keypoints and texture prediction.

\subsection{Experimental Setup}
\seclabel{expsetup}
\noindent \textbf{Dataset.} We use the CUB-200-2011 dataset \cite{Wah}, which has 6000 training and test images of 200 species of birds. Each image is annotated with the bounding box, visibility indicator and locations of 14 semantic keypoints, and the ground truth foreground mask. We filter out nearly 300 images where the visible number of keypoints are less than or equal to 6, since these typically correspond to truncated close shots. We divide the test set in half to create a validation set, which we use for hyper-parameter tuning.
\begin{figure*}[p!]
  \centering
\includegraphics[width=\textwidth]{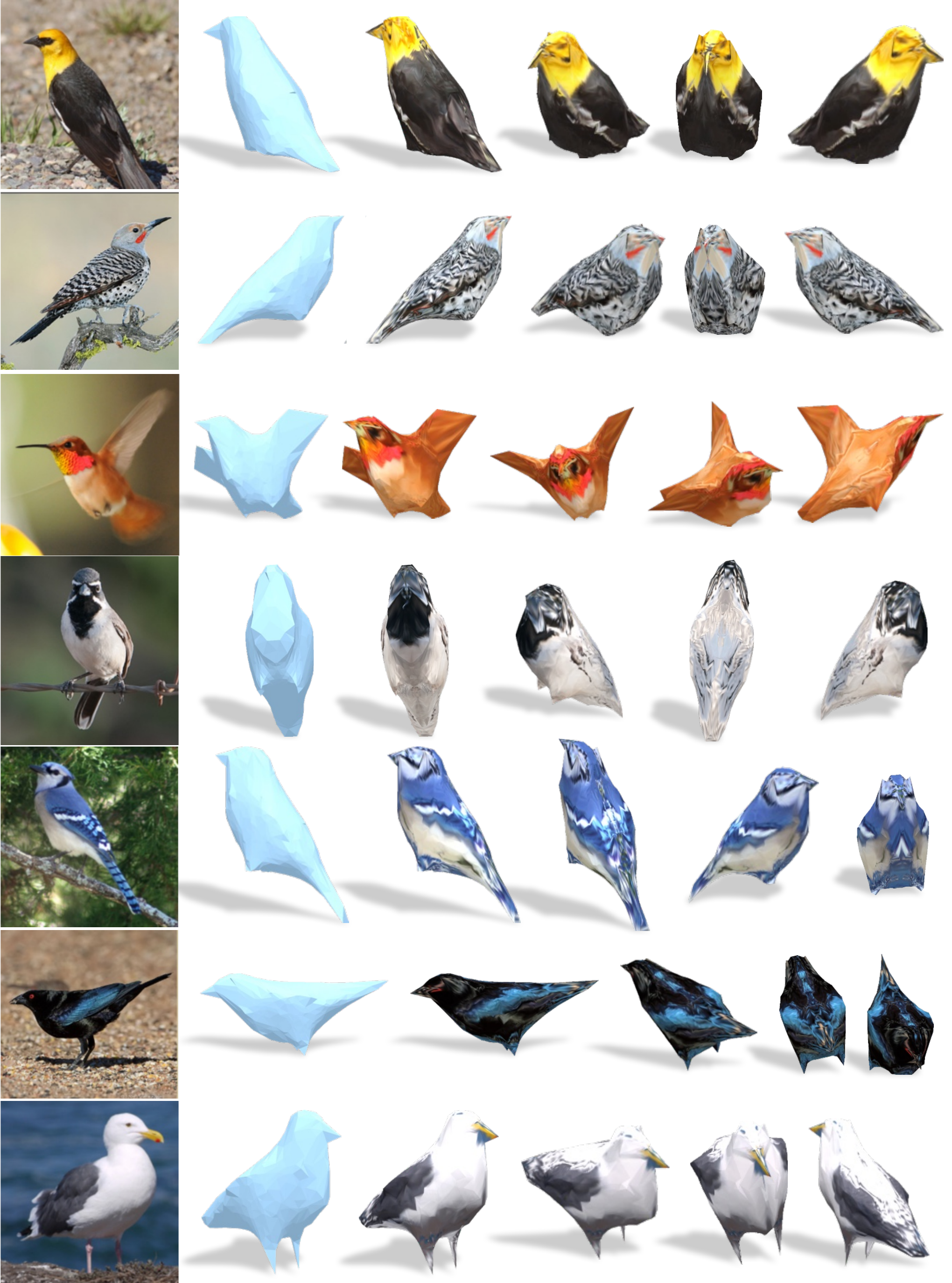}
  \caption{\small {\bf Sample results.} We show predictions of our approach on
    images from the test set. For each input image on the left, we visualize (in
    order): the predicted 3D shape and texture viewed from the predicted camera,
    and textured shape from three novel viewpoints. See the appendix for additional randomly selected results and video at \url{https://akanazawa.github.io/cmr/}.
  }
  \figlabel{big_figure}
\end{figure*}

\noindent \paragraph{\emph{\textbf{Network Architecture.}}}
A schematic of the various modules of our prediction network is depicted in \figref{overview}. The encoder consists of an ImageNet pretrained ResNet-18~\cite{He2016}, followed by a convolutional layer that downsamples the spatial and the channel dimensions by half. This is vectorized to form a 4096-D vector, which is sent to two fully-connected layers to get to the shared latent space of size 200. The deformation and the camera prediction components are linear layers on top of this latent space. The texture flow component consists of 5 upconvolution layers where the final output is passed through a $tanh$ function to keep the flow in a normalized [-1, 1] space.
We use the neural mesh renderer~\cite{NMR} so all rendering procedures are differentiable. All images are cropped using the instance bounding box and resized such that the maximum image dimension is 256. We augment the training data on the fly by jittering the scale and translation of the bounding box and with image mirroring. Our mesh geometry corresponds to that of a perfectly symmetric sphere with 642 vertices and 1280 faces.

\subsection{Qualitative Results}
\seclabel{qualitative}
We visualize the results and application of our learned predictor using the CUB dataset. We show various reconstructions corresponding to different input images, visualize some of the deformation modes learned, and show that the common deformable model parametrization allows us to transfer the texture of one instance onto another.

\noindent \paragraph{\emph{\textbf{Single-view 3D Reconstruction.}}}
We show sample reconstruction results on images from the CUB test set in \figref{big_figure}. We show the predicted shape and texture from the inferred camera viewpoint, as well as from novel views. Please see appendix for additional randomly selected samples and videos showing the results from 360 views.

We observe that our learned model can accurately predict the shape, estimate the
camera and also infer meaningful texture from the corresponding input image. Our
predicted 3D shape captures the overall shape (fat or thin birds), and even some
finer details \eg beaks or large deformations \eg flying birds. Additionally,
our learned pose and texture prediction are accurate and realistic across
different instances. We observe that the error modes corresponds to not
predicting rare poses,  and inability to incorporate asymmetric
articulation. However, we feel that these predictions learned using only an
annotated image collection are encouraging.

\setlength{\columnsep}{10pt}
\setlength{\intextsep}{5pt}
\begin{wrapfigure}{r}{0.46\textwidth}
  \centering
  \includegraphics[width=.9\textwidth]{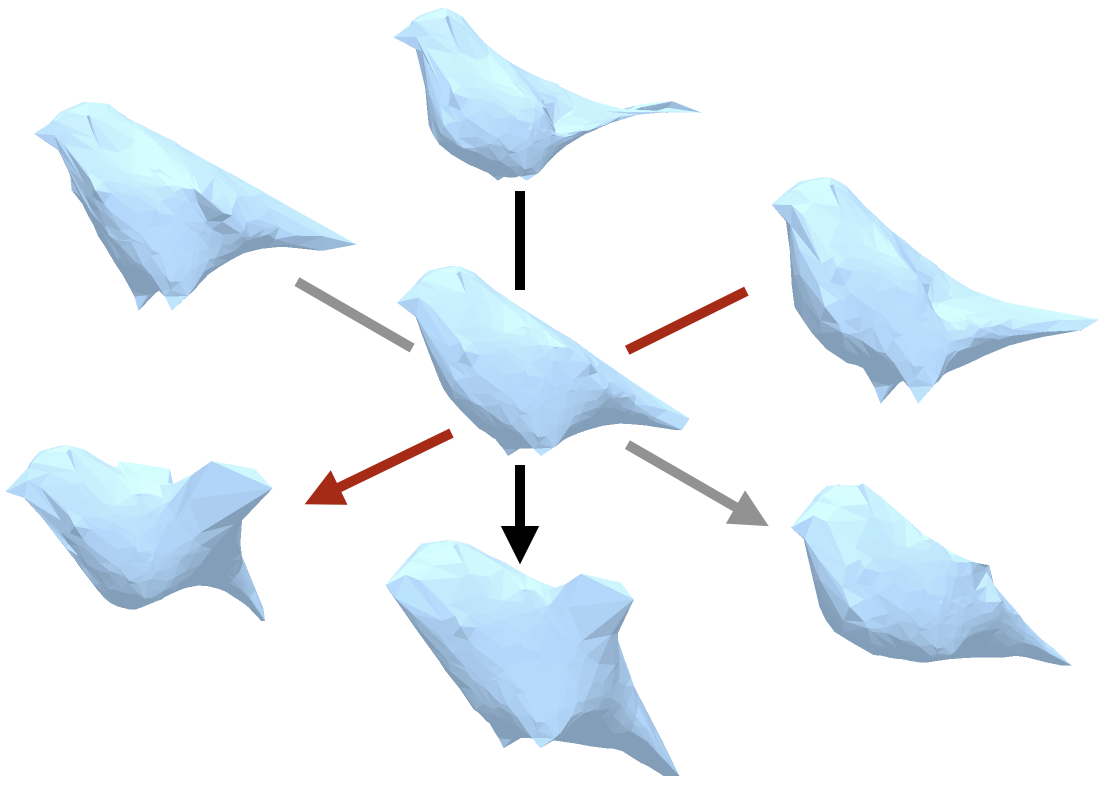}
  \caption{\small {\bf Learned deformation modes.} We visualize the space of learned shapes by depicting the mean shape (centre) and three common modes of deformation as obtained by PCA on the predicted deformations across the dataset.}
  \figlabel{pca}
\end{wrapfigure}

\noindent \paragraph{\emph{\textbf{Learned shape space.}}}
The presented approach represents the shape of an instance via a category-level learned mean shape and a per-instance predicted deformation $\Delta_V$. To gain insight into the common modes of deformation captured via our predictor, obtained the principal deformation modes by computing PCA on the predicted deformations across all instances in the training set.

We visualize in \figref{pca} our mean shape deformed in directions corresponding three common deformation modes. We note that these plausibly correspond to some of the natural factors of variation in the 3D structure across birds \eg fat or thin birds, opening of wings, deformation of tails and legs.

\noindent \paragraph{\emph{\textbf{Texture Transfer.}}} Recall that the textures of different instance in our formulation are captured in a canonical appearance space in the form of a predicted `texture image' $I_{uv}$. This parametrization allows us to easily modify the surface appearance, and in particular transfer texture across instances.

We show some results in \figref{texture_transfer} where we sample pairs of instances, and transfer the texture from one image onto the predicted shape of the other. We can achieve this by simply using the predicted texture image corresponding to the first when rendering the predicted 3D for the other. We note that even though the two views might be different, since the underlying `texture image' space is consistent, the transferred texture is also semantically consistent \eg the colors corresponding to the one bird's body are transferred onto the other bird's body.

\begin{figure*}[t]
  \centering
 \includegraphics[width=\textwidth]{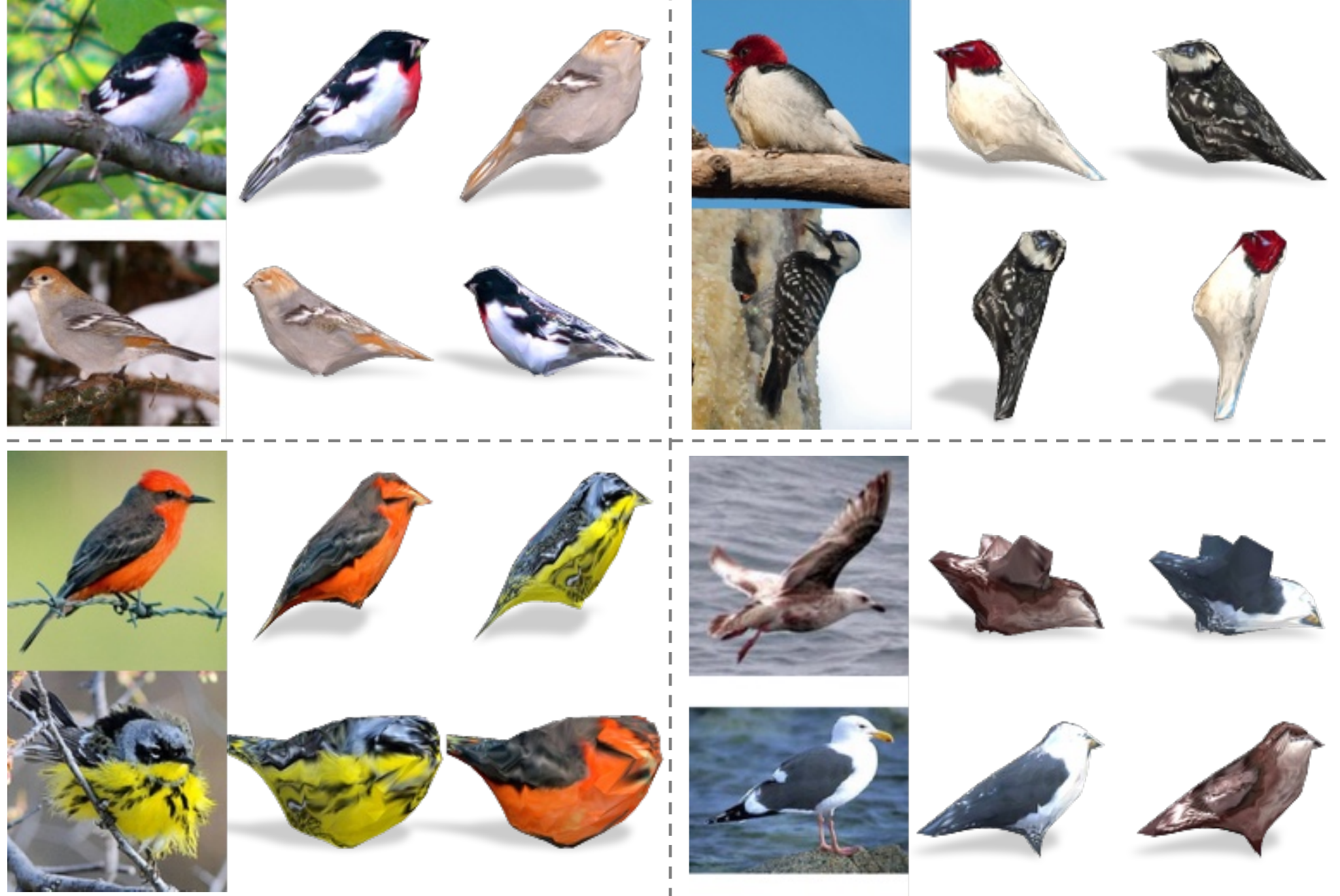}
  \caption{\small {\bf Texture Transfer Results.} Our representation allows us to easily transfer the predicted texture across instances using the canonical appearance image (see text for details). We visualize sample results of texture transfer across different pairs of birds. For each pair, we show (left): the input image, (middle): the predicted textured mesh from the predicted viewpoint, and (right): the predicted mesh textured using the predicted texture of the other bird.}
  \figlabel{texture_transfer}
\end{figure*}

\subsection{Quantitative Evaluation}
\seclabel{quantitative}
We attempt to indirectly measure the quality of our recovered reconstructions on the CUB dataset. As there is no ground-truth 3D available for benchmarking, we instead evaluate the mask reprojection accuracy. For each test instance in the CUB dataset, we obtain a mask prediction via rendering the predicted 3D shape from the predicted  camera viewpoint. We then compute the intersection over union (IoU) of this predicted mask with the annotated ground-truth mask. Note that to correctly predict the foreground mask, we need both, accurate shape and accurate camera.

Our results are plotted in \figref{maskiou}. We compare the accuracy our full shape prediction (using learned mean shape $\bar{V}$ and predicted deformation $\Delta_V$) against only using the learned mean shape to obtain the predicted mask. We observe that the predicted deformations result in improvements, indicating that we are able to capture the specifics of the shape of different instances. Additionally, we also report the performance using the camera obtained via structure from motion (which uses ground-truth annotated keypoints) instead of using the predicted camera. We note that comparable results in the two settings demonstrate the accuracy of our learned camera estimation. Lastly, we can also measure our keypoint reprojection accuracy using the percentage of correct keypoints (PCK) metric \cite{Yang}. We similarly observe that our full predicted shape performs (slightly) better than only relying on the category-level mean shape -- by obtaining a PCK (at normalized distance threshold 0.1) of 0.81 compared to  0.80. The improvement over the mean shape is less prominent in this scenario as most of the semantic keypoints defined are on the torso and therefore typically undergo only small deformations.

\renewcommand{\arraystretch}{1.2}
\setlength{\tabcolsep}{2pt}

\begin{figure}
\TopFloatBoxes
\begin{floatrow}
\ffigbox[1.2\FBwidth]{%
  \includegraphics[width=.45\textwidth]{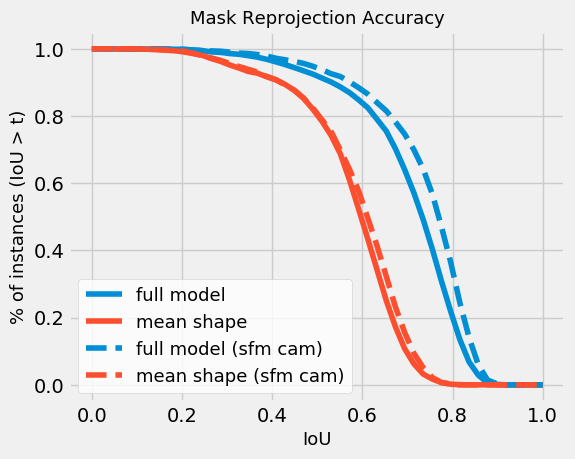}
}{%
  \caption{\small {\bf Mask reprojection accuracy evaluation on CUB.} We plot the fraction of test instances with IoU between the predicted and ground-truth mask higher than different thresholds (higher is better) and compare the predictions using the full model against only using the learned mean shape. We report the reprojection accuracy using predicted cameras and cameras obtained via structure-from-motion based on keypoint annotation.}
  \figlabel{maskiou}%
}
\capbtabbox{%
    \begin{tabular}{l c c}
    \toprule
    Method & Aeroplane & Car  \\ \midrule
    CSDM~\cite{CSDM}   & 0.40 & 0.60 \\
    DRC~\cite{drcTulsiani17}    & 0.42 & 0.67 \\
    Ours   & 0.46 & 0.64 \\ \bottomrule
    \end{tabular}
}{%
  \caption{\small {\bf Reconstruction evaluation using PASCAL 3D+.} We report the mean intersection over union (IoU) on PASCAL 3D+ to benchmark the obtained 3D reconstructions (higher is better). We compare to previous deformable model fitting-based~\cite{CSDM} and volumetric prediction ~\cite{drcTulsiani17} approaches that use similar image collection supervision. Note that our approach can additionally predict texture and semantics.}
  \tablelabel{p3d}
}
\end{floatrow}
\end{figure}

\subsection{Evaluation on Other Object Classes}
\seclabel{pascal3d}


\begin{figure*}[t]
  \centering
  \includegraphics[width=\textwidth]{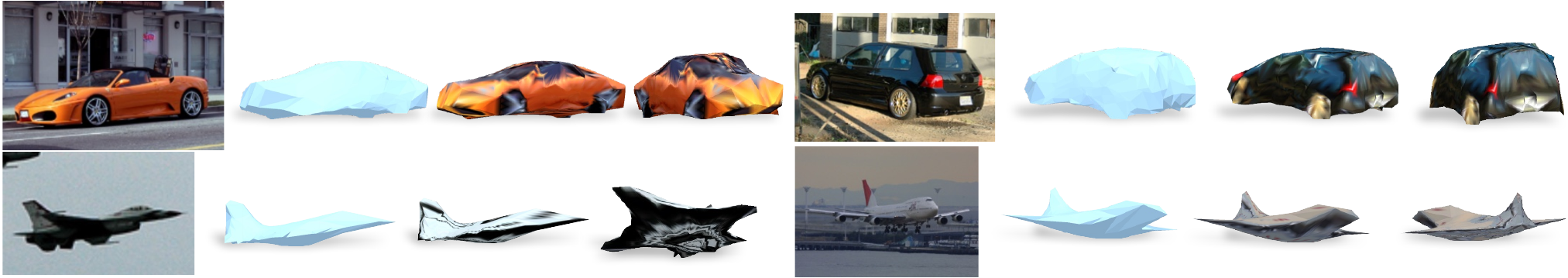}
  \caption{\small {\bf Pascal 3D+ results.} We show predictions of our approach on images from the test set. For each input image on the left, we visualize (in order): the predicted 3D shape viewed from the predicted camera, the predicted shape with texture viewed from the predicted camera, and the shape with texture viewed from a novel viewpoint.}
  \figlabel{p3dresults}
\end{figure*}
While our primary results focus on predicting the 3D shape and texture of birds using the CUB dataset, we note that some previous approaches have examined the task of shape inference/prediction using a similar annotated image collection as supervision.  While these previous methods do not infer texture, we can compare our shape predictions against those obtained by these techniques. 

We compare to previous deformable model fitting-based~\cite{CSDM} and volumetric prediction ~\cite{drcTulsiani17} methods using the PASCAL 3D+ dataset and examine the car and aeroplane categories. Both of these approaches can leverage the annotation we have available \ie segmentation masks and keypoints to learn 3D shape inference (although \cite{drcTulsiani17} requires annotated cameras instead of keypoints). Similar to \cite{drcTulsiani17}, we use PASCAL VOC and Imagenet images with available keypoint annotations from PASCAL3D+ to train our model, and use an off-the shelf segmentation algorithm~\cite{maskRCNN} to obtain foreground masks for the ImageNet subset.

We report the mean IoU evaluation on the test set in \tableref{p3d} and observe that we perform comparably, if not better than these alternate methods. We also note that our approach yields additional outputs \eg texture, that these methods do not. We visualize some predictions in \figref{p3dresults}. While our predicted shapes are often reasonable, the textures have more errors due to shiny regions (\eg for cars) or smaller amount of training data (\eg for aeroplanes).

\section{Discussion}
We have presented a framework for learning single-view prediction of a textured 3D mesh using an image collection as supervision. While our results represent an encouraging step, we have by no means solved the problem in the general case, and a number of interesting challenges and possible directions remain. Our formulation addresses shape change and articulation via a similar shape deformation mechanism, and it may be beneficial to extend our deformable shape model to explicitly allow articulation. Additionally, while we presented a method to synthesize texture via copying image pixels, a more sophisticated mechanism that allows both, copying image content and synthesizing novel aspects might be desirable. Finally, even though we can learn using only a single-view per training instance, our approach may be equally applicable, and might yield perhaps even better results, for the scenario where multiple views per training instance are available. However, on the other end of the supervision spectrum, it would be desirable to relax the need of annotation even further, and investigate learning similar prediction models using unannotated image collections.

\paragraph{\emph{\textbf{Acknowledgements.}}}
{We thank David Fouhey for the creative title suggestions and members of the BAIR community for helpful discussions and comments. This work was supported in part by Intel/NSF VEC award IIS-1539099, NSF Award IIS-1212798, and BAIR sponsors.}

\clearpage
\bibliographystyle{splncs04}
\bibliography{references}

\clearpage
\section*{Appendix}
\subsection*{A1. Optimization Details}
\textbf{Mesh Geometry.}
The geometry of predicted mesh corresponds to that of an `Icoshpere' (subdivided Icosahedron) at subdivision level 3 (see ~\cite{icosphere} for an excellent description and implementation). This results in a mesh with 642 vertices $V$ and 1280 faces $F$. We keep the faces fixed during our learning process, and predict (via a learned mean shape and predicted deformations), the positions of the vertices $V$.

\paragraph{\emph{\textbf{Symmetry.}}}
We enforce reflectional symmetry along the X-axis. As the initial icosphere is perfectly symmetric, each vertex $v \in V$ either lies on the $YZ$ plane, or has a corresponding symmetric vertex. For each pair of symmetric vertices, say $(v_i, v_j)$, we only treat the location of one vertex in the pair (say $v_i$) as a free parameter. As a consequence, we predict the location of $337$ vertices (32 of these are on the $YZ$ plane, and 305 from one symmetric vertex pair each) to instantiate the mesh vertex locations $V$.

\paragraph{\emph{\textbf{Mean Shape Initialization.}}}
While the initial icosphere yields some default positions for the vertices $V$,
following previous approaches~\cite{Cashman,CSDM}, we find it beneficial to
instead use a better initialization for the vertex locations $\bar{V}$ in the
mean shape. To this end, we use the convex hull of the mean keypoint locations
obtained after running structure-from-motion using the annotated keypoints. For
each of the vertices $v \in \bar{V}$, its initial position is computed by
projecting it onto this convex hull. Our learning process therefore starts with
this coarse convex-hull mean shape initialization, and over the course of the
training, learns a better mean shape. The initial and the final meanshapes are
illustrated in \figref{meanshape}.
\begin{figure}[h]
  \centering
  \includegraphics[width=\textwidth]{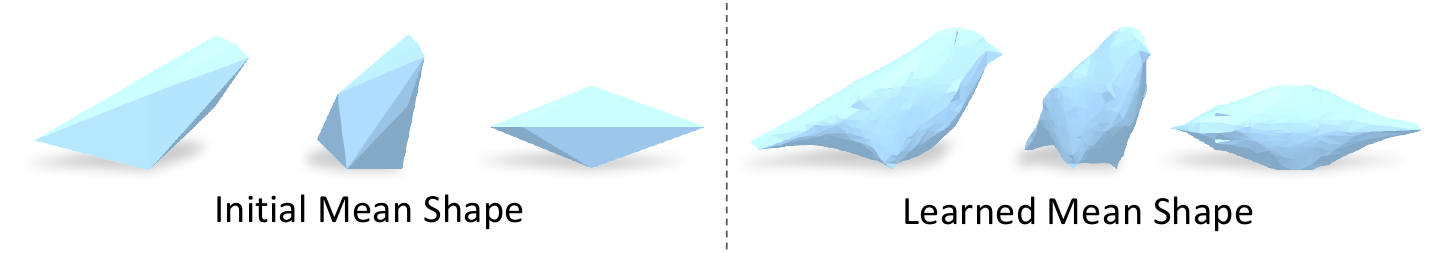}
  \caption{\small {\bf Initial and Learned Mean Shapes.} On the left we show the initial mean shape obtained from running SfM on the annotated keypoints. We use this as initialization. On the right we show the final learned mean shape. 
  }
  \figlabel{meanshape}
\end{figure}

\paragraph{\emph{\textbf{Laplacian Smoothness.}}}
As a prior for smoothness, we minimize the mean mesh curvature. The curvature at the vertices can be computed via a discretization of the continuous Laplace-Beltrami operator. Assuming $N$ vertices, this (discrete) operator (denoted as the laplacian $L$) is simply a fixed $N \times N$ sparse matrix, and the matrix $LV$ yields the normal direction at each vertex weighted by the curvature. We can therefore minimize the mean norm of the rows of $LV$ to minimize mean curvature. We use the `cotangent weights'~\cite{pinkall1993computing} to define $L$, as it accounts for the local geometry instead of just adjacency. We refer the reader to Section 2.1 of \cite{sorkine2006differential} for a concise review of the concepts involved.


\subsection*{A2. Additional Results and Comparisons}

\paragraph{\emph{\textbf{Randomly selected results}}}
In \figref{supp1} and \figref{supp2} we show predictions of our approach on 40
\emph{randomly selected} images from the test set. In each column, we show the
input image followed by the predicted 3D shape and texture from the predicted
camera view, and three views of the textured shape corresponding to a rotation
of 60, 180 and -60 degrees around y-axis.

\paragraph{\emph{\textbf{Comparison with directly sampled texture}}}
In \figref{text_comp}, we compare our texture predictions with an alternative approach of directly copying the textures from the image. Given the predicted camera and mesh, we paint onto the inferred shape by directly sampling the pixel values of the visible regions with the symmetric texture map. We show the input image, the results from predicted camera view, and three different viewpoints. As seen in the figure, the direct sampling approach results in holes (regions in magenta) and includes background pixels, which results in unnatural texture when seen from novel views (\eg example two, right most column).

\begin{figure*}[p]
  \centering
  \includegraphics[width=\textwidth]{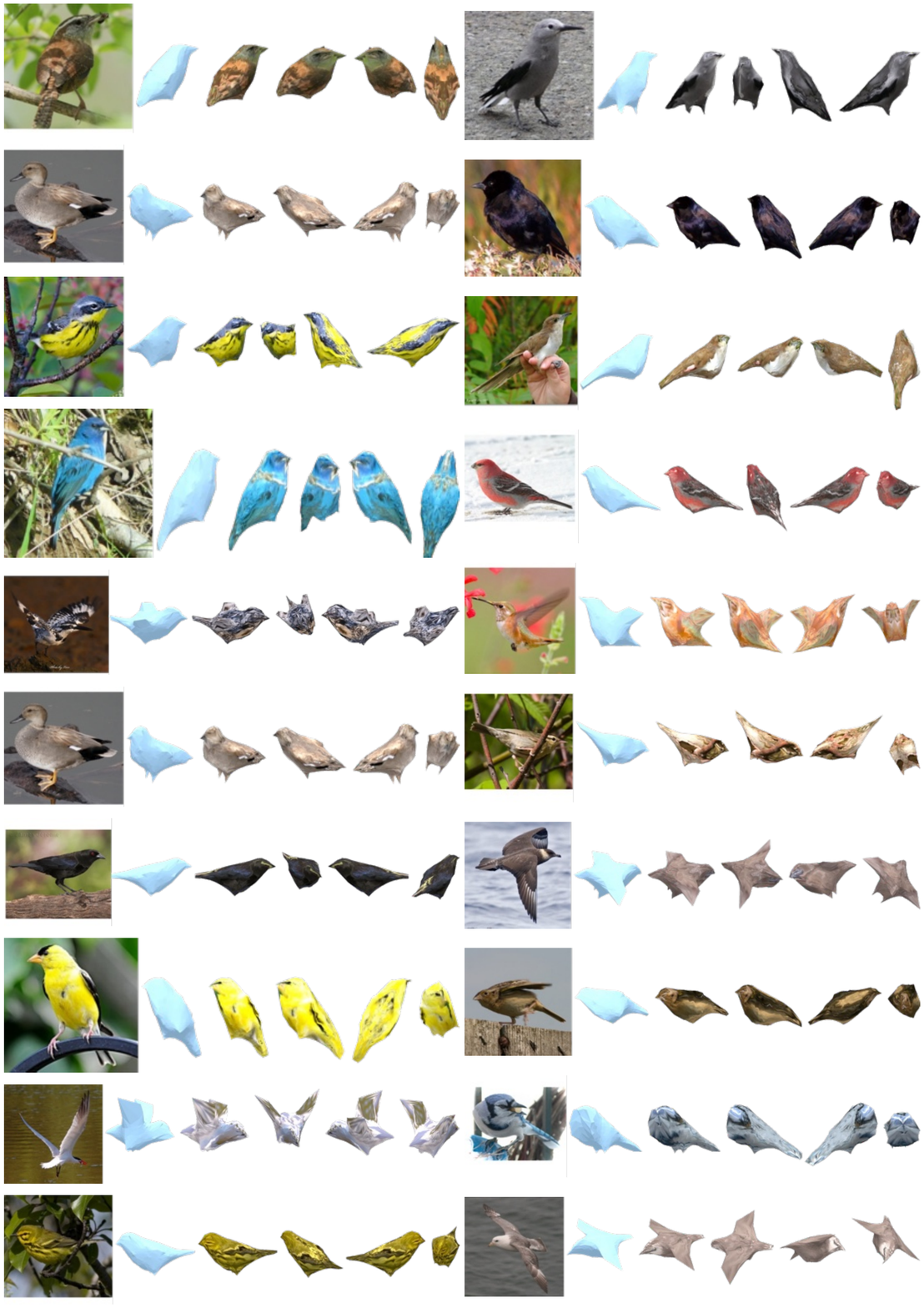}
  \caption{\small {\bf Randomly selected results.} We show predictions of our approach on \emph{random
    images} from the test set. For each column, we show the input image on the left and visualize (in
    order): the predicted 3D shape and texture viewed from the predicted camera,
    and textured shape from three novel views corresponding to a rotation of 60, 180 and -60 degrees around y-axis.
  }
  \figlabel{supp1}
\end{figure*}

\begin{figure*}[p]
  \centering
  \includegraphics[width=\textwidth]{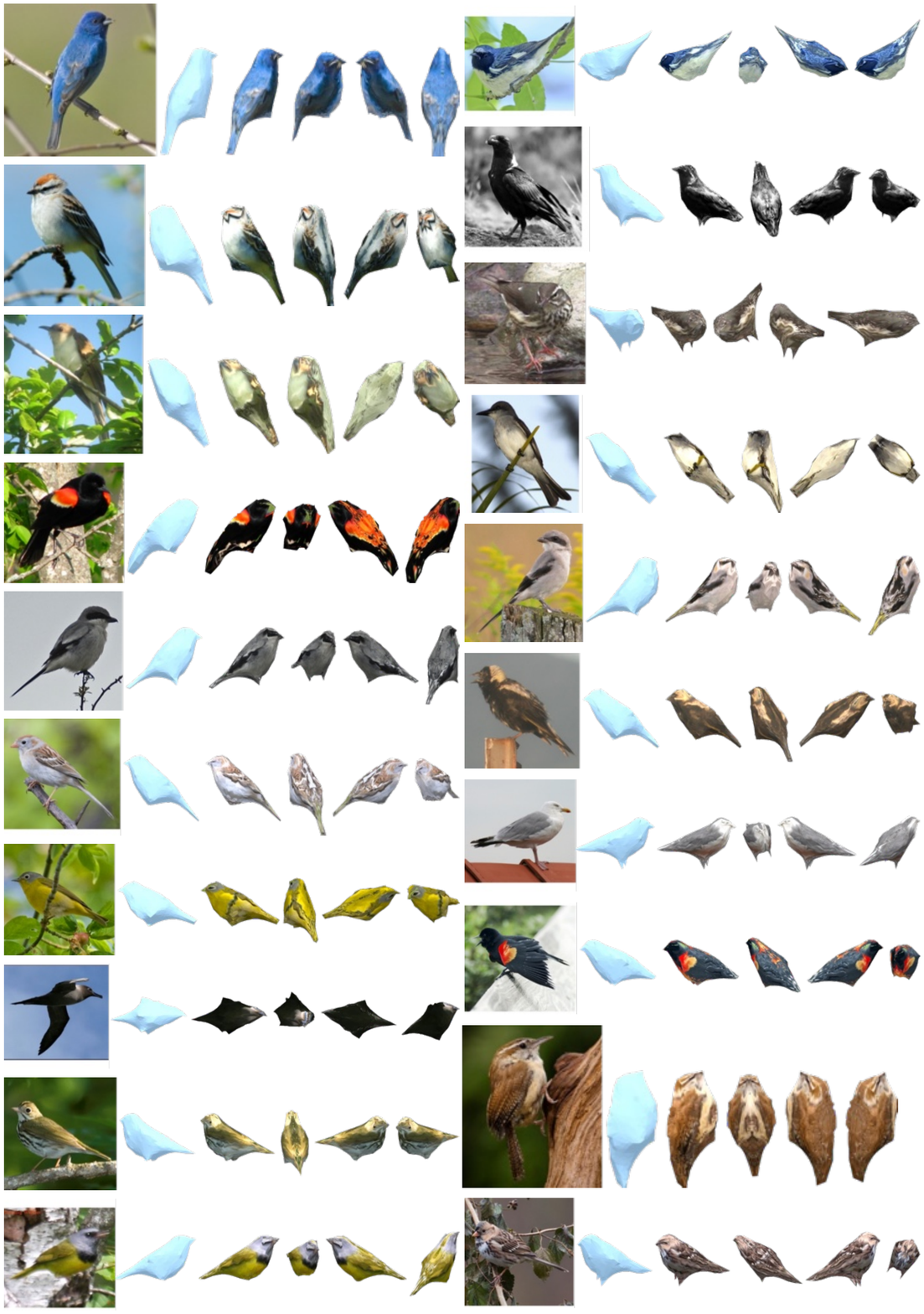}
  \caption{\small {\bf Randomly selected results.} We show predictions of our approach on \emph{random
    images} from the test set. For each column, we show the input image on the left and visualize (in
    order): the predicted 3D shape and texture viewed from the predicted camera,
    and textured shape from three novel views corresponding to a rotation of 60, 180 and -60 degrees around y-axis.
  }
  \figlabel{supp2}
\end{figure*}

\begin{figure*}[p]
  \centering
  \includegraphics[width=\textwidth]{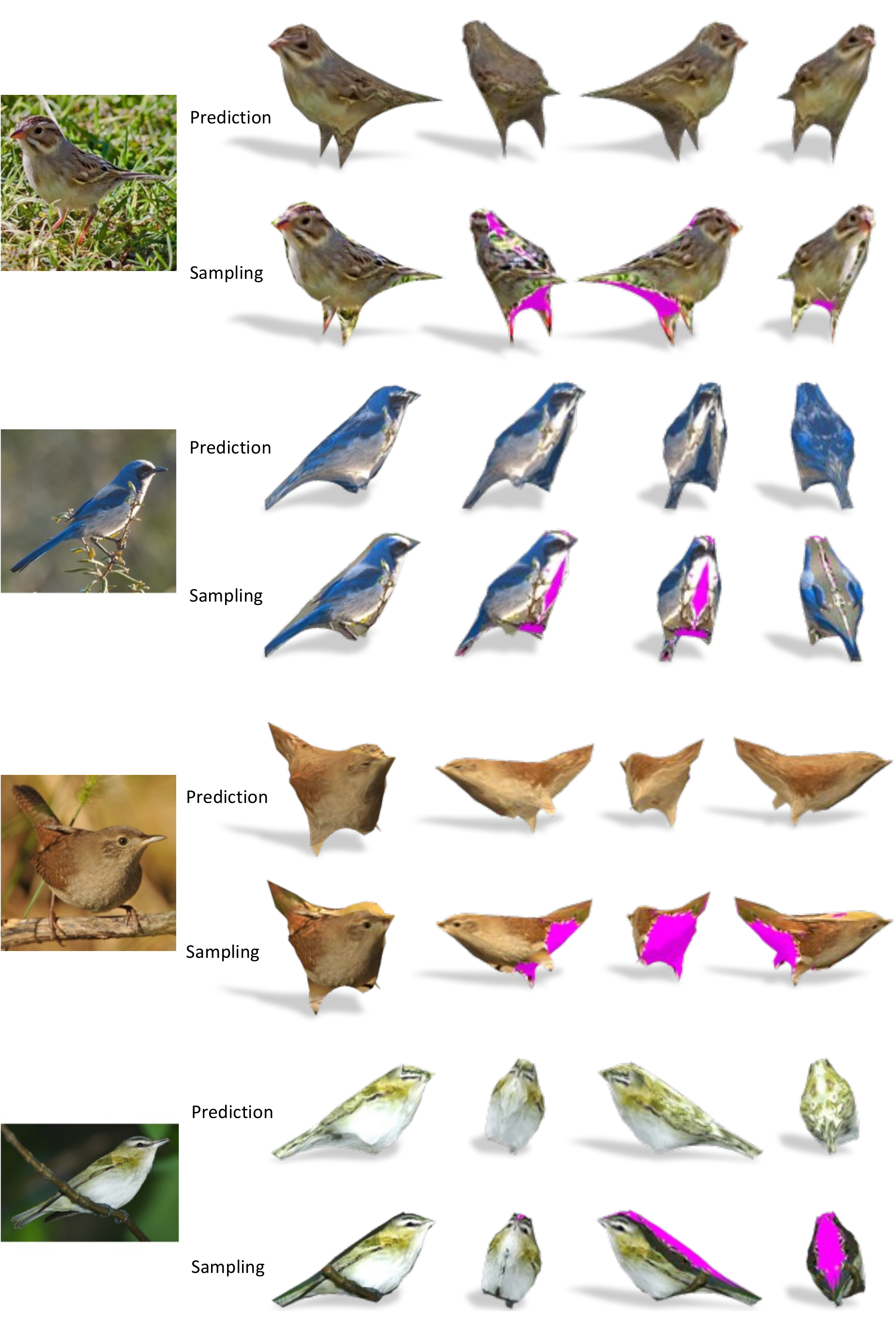}
  \caption{\small {\bf Inferred vs sampled texture.} We compare our texture predictions with a baseline approach of directly copying the pixel values that the predicted mesh and camera projects onto. We show the input image, results from the predicted camera and three other viewpoints. Note that sampling approach, even after symmetrizing the visible region, results in holes in the texture map (shown in magenta) and includes background pixels.}
  \figlabel{text_comp}
\end{figure*}

\end{document}